\documentclass[runningheads]{llncs}
\usepackage[T1]{fontenc}
\usepackage{graphicx}
\usepackage{hyperref}
\usepackage{color}

\usepackage{amsfonts,amssymb} 
\usepackage{amsmath}
\usepackage{booktabs,multirow}
\usepackage{xcolor}
\usepackage[misc]{ifsym}

\def\ie{\emph{i.e.}}
\def\eg{\emph{e.g.}}

\begin{document}
\title{CellGAN: Conditional Cervical Cell Synthesis for Augmenting Cytopathological Image Classification}
\titlerunning{CellGAN: Conditional Cervical Cell Synthesis}

\author{
Zhenrong Shen\inst{1} \and
Maosong Cao\inst{2} \and
Sheng Wang\inst{1,3} \and
Lichi Zhang\inst{1} \and \\
Qian Wang\inst{2}$^{(\textrm{\Letter})}$
}
\authorrunning{Z. Shen et al.}

\institute{
School of Biomedical Engineering, Shanghai Jiao Tong University, Shanghai, China \and
School of Biomedical Engineering, ShanghaiTech University, Shanghai, China
\email{wangqian2@shanghaitech.edu.cn} \and
Shanghai United Imaging Intelligence Co., Ltd., Shanghai, China
}
\maketitle              
\begin{abstract}
Automatic examination of thin-prep cytologic test (TCT) slides can assist pathologists in finding cervical abnormality for accurate and efficient cancer screening. 
Current solutions mostly need to localize suspicious cells and classify abnormality based on local patches, concerning the fact that whole slide images of TCT are extremely large.
It thus requires many annotations of normal and abnormal cervical cells, to supervise the training of the patch-level classifier for promising performance. 
In this paper, we propose CellGAN to synthesize cytopathological images of various cervical cell types for augmenting patch-level cell classification. 
Built upon a lightweight backbone, CellGAN is equipped with a non-linear class mapping network to effectively incorporate cell type information into image generation.
We also propose the Skip-layer Global Context module to model the complex spatial relationship of the cells, and attain high fidelity of the synthesized images through adversarial learning.
Our experiments demonstrate that CellGAN can produce visually plausible TCT cytopathological images for different cell types. 
We also validate the effectiveness of using CellGAN to greatly augment patch-level cell classification performance.
Our code and model checkpoint are available at \url{https://github.com/ZhenrongShen/CellGAN}.

\keywords{Conditional Image Synthesis \and Generative Adversarial Network \and Cytopathological Image Classification \and Data Augmentation.}
\end{abstract}

\section{Introduction}
Cervical cancer accounts for 6.6\% of the total cancer deaths in females worldwide, making it a global threat to healthcare~\cite{gultekin2020world}. 
Early cytology screening is highly effective for the prevention and timely treatment of cervical cancer~\cite{patel2011cervical}. 
Nowadays, thin-prep cytologic test (TCT)~\cite{abulafia2003performance} is widely used to screen cervical cancers according to the Bethesda system (TBS) rules~\cite{nayar2015bethesda}. 
Typically there are five types of cervical squamous cells under TCT examinations~\cite{davey1994atypical}, including 
normal class or negative for intraepithelial malignancy (NILM), 
atypical squamous cells of undetermined significance (ASC-US),
low-grade squamous intraepithelial lesion (LSIL), 
atypical squamous cells that cannot exclude HSIL (ASC-H), 
and high-grade squamous intraepithelial lesion (HSIL). 
The NILM cells have no cytological abnormalities while the others are manifestations of cervical abnormality to a different extent. 
By observing cellular features (\eg, nucleus-cytoplasm ratio) and judging cell types, pathologists can provide a diagnosis that is critical to the clinical management of cervical abnormality.

After scanning whole-slide images (WSIs) from TCT samples, automatic TCT screening is highly desired due to the large population versus the limited number of pathologists.
As the WSI data per sample has a huge size, the idea of identifying abnormal cells in a hierarchical manner has been proposed and investigated by several studies using deep learning~\cite{xiang2020novel,zhou2021hierarchical,cao2021novel}.
In general, these solutions start with the extraction of suspicious cell patches and then conduct patch-level classification. 
The promising performance of cell classification at the patch level is critical, which contributes to sample-level diagnosis after integrating outcomes from many patches in a WSI.
However, such a patch-level classification task requires a large number of annotated training data. 
And the efforts in collecting reliably annotated data can hardly be negligible, which requires high expertise due to the intrinsic difficulty of visually reading WSIs. 

To alleviate the shortage of sufficient data to supervise classification, one may adopt traditional data augmentation techniques, which yet may bring little improvement due to scarcely expanded data diversity~\cite{shorten2019survey}. 
Thus, synthesizing cytopathological images for cervical cells is highly desired to effectively augment training data.
Existing literature on pathological image synthesis has explored the generation of histopathological images~\cite{hou2019robust,xue2021selective}. 
In cytopathological images, on the contrary, cervical cells can be spatially isolated from each other, or are highly squeezed and even overlapped. 
The spatial relationship of individual cells is complex, adding diversity to the image appearance of color, morphology, texture, etc.
In addition, the differences between cell types are mainly related to nuanced cellular attributes, thus requiring fine granularity in modulating synthesized images toward the expected cell types.
Therefore, the task to synthesize realistic cytopathological images becomes very challenging.

\begin{figure}[ht]
    \centering
    \includegraphics[width=\textwidth]{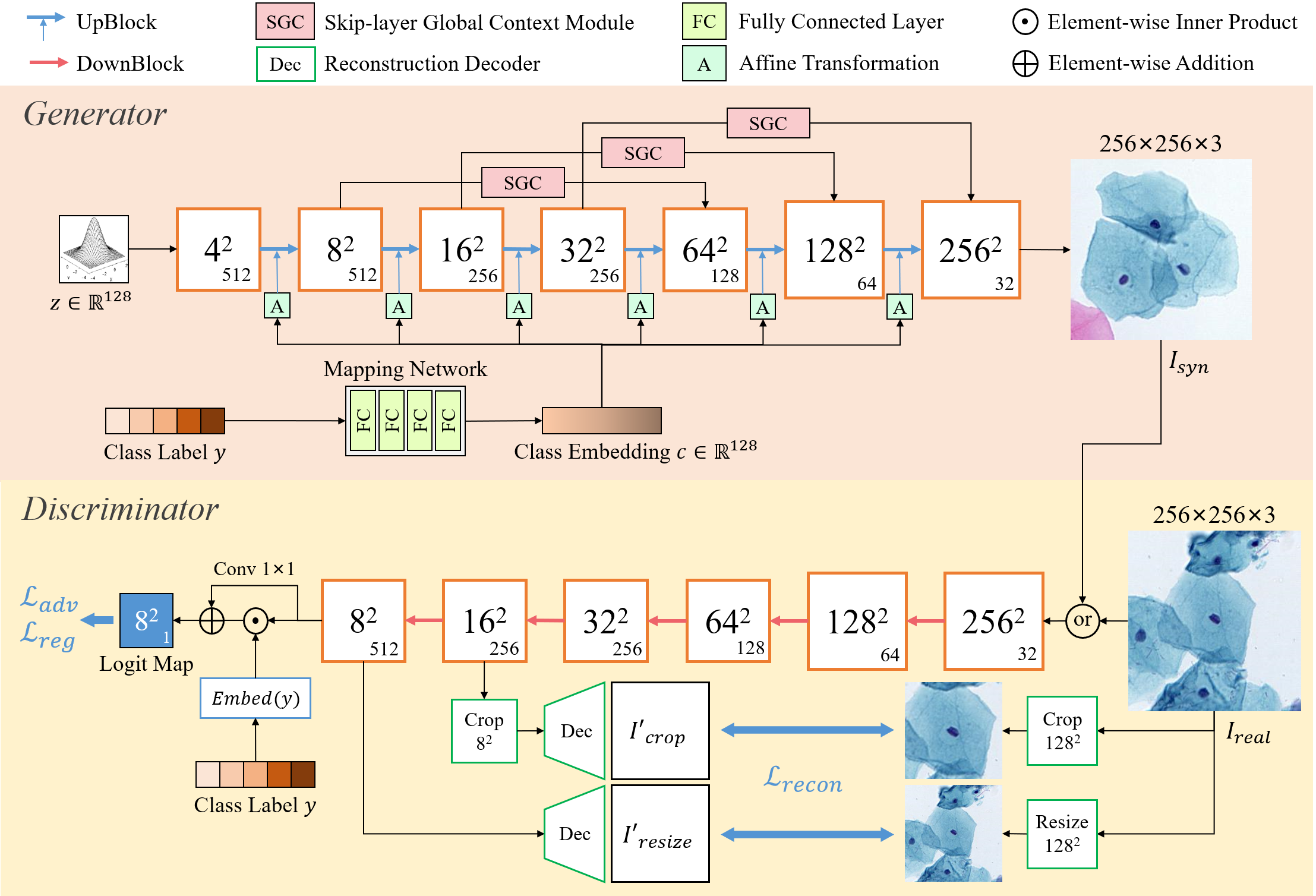}
    \caption{Overall architecture of the proposed CellGAN. The numbers in the center and the bottom right corner of each square indicate the feature map size and the channel number, respectively.}
    \label{fig:overview}
\end{figure}

Aiming at augmenting the performance of cervical abnormality screening, we develop a novel conditional generative adversarial network in this paper, namely CellGAN, to synthesize cytopathological images for various cell types. 
We leverage FastGAN~\cite{liu2020towards} as the backbone for the sake of training stability and computational efficiency.
To inject cell type for fine-grained conditioning, a non-linear mapping network embeds the class labels to perform layer-wise feature modulation in the generator.
Meanwhile, we introduce the Skip-layer Global Context (SGC) module to capture the long-range dependency of cells for precisely modeling their spatial relationship.
We adopt an adversarial learning scheme, where the discriminator is modified in a projection-based way~\cite{miyato2018cgans} for matching conditional data distribution.
To the best of our knowledge, our proposed CellGAN is the first generative model with the capability to synthesize realistic cytopathological images for various cervical cell types. 
The experimental results validate the visual plausibility of CellGAN synthesized images, as well as demonstrate their data augmentation effectiveness on patch-level cell classification.

\section{Method}
The dilemma of medical image synthesis lies in the conflict between the limited availability of medical image data and the high demand for data amount to train reliable generative models.
To ensure the synthesized image quality given relatively limited training samples, the proposed CellGAN is built upon FastGAN~\cite{liu2020towards} towards stabilized and fast training for few-shot image synthesis.
By working in a class-conditional manner, CellGAN can explicitly control the cervical squamous cell types in the synthesized cytopathological images, which is critical to augment the downstream classification task.
The overall architecture of CellGAN is presented in Fig.~\ref{fig:overview}, and more detailed structures of the key components are displayed in Supplementary Materials.


\subsection{Architecture of the Generator} 
The generator of CellGAN has two input vectors. 
The first input of the class label $y$, which adopts one-hot encoding, provides class-conditional information to indicate the expected cervical cell type in the synthesized image $I_{syn}$.
The second input of the 128-dimensional latent vector $z$ represents the remaining image information, from which $I_{syn}$ is gradually expanded.
We stack six UpBlocks to form the main branch of the generator.

To inject cell class label $y$ into each UpBlock, we follow a similar design to StyleGAN~\cite{karras2019style}.
Specifically, the class label $y$ is first projected to a class embedding $c$ via a non-linear mapping network, which is implemented using four groups of fully connected layers and LeakyReLU activations.
We set the dimensions of class embedding $c$ to the same as the latent vector $z$.
Then, we pass $c$ through learnable affine transformations, such that the class embedding is specialized to the scaling and bias parameters controlling Adaptive Instance Normalization (AdaIN)~\cite{karras2019style} in each UpBlock. 
The motivation for the design above comes from our hypothesis that the class-conditional information mainly encodes cellular attributes related to cell types, rather than common image appearance. 
Therefore, by modulating the feature maps at multiple scales, the input class label can better control the generation of cellular attributes. 

We further introduce the Skip-layer Global Context (SGC) module into the generator (see Fig.2 in Supplementary Materials), to better handle the diversity of the spatial relationship of the cells. 
Our SGC module reformulates the idea of GCNet~\cite{cao2019gcnet} with the design of SLE module from FastGAN~\cite{liu2020towards}.
It first performs global context modeling on the low-resolution feature maps, then transforms global context to capture channel-wise dependency, and finally merges the transformed features into high-resolution feature maps.
In this way, the proposed SGC module learns a global understanding of the cell-to-cell spatial relationship and injects it into image generation via computationally efficient modeling of long-range dependency.

\subsection{Discriminator and Adversarial Training}
In an adversarial training setting, the discriminator forces the generator to faithfully match the conditional data distribution of real cervical cytopathological images, thus prompting the generator to produce visually and semantically realistic images.
For training stability, the discriminator is trained as a feature encoder with two extra decoders. 
In particular, five ResNet-like~\cite{he2016deep} DownBlocks are employed to convert the input image into an $8\times8\times512$ feature map. 
Two simple decoders reconstruct downscaled and randomly cropped versions of input images $I'_{crop}$ and $I'_{resize}$ from $8^{2}$ and $16^{2}$ feature maps, respectively.
These decoders are optimized together with the discriminator by using a reconstruction loss $\mathcal{L}_{recon}$ that is represented below:

\begin{equation}
    \mathcal{L}_{recon}=\mathbb{E}_{f\sim Dis(x),x\sim I_{real}}\left[{\left\|Dec(f)-\mathcal{T}(x)\right\|}_{\ell_1}\right],
\end{equation}

\noindent where $\mathcal{T}$ denotes the image processing (\ie, $\frac{1}{2}$ downsampling and $\frac{1}{4}$ random cropping) on real image $I_{real}$,
$f$ is the processed intermediate feature map from the discriminator $Dis$, 
and $Dec$ stands for the reconstruction decoder.
This simple self-supervised technique provides a strong regularization in forcing the discriminator to extract a good image representation.

To provide more detailed feedback from the discriminator, PatchGAN~\cite{isola2017image} architecture is adopted to output an $8\times8$ logit map by using a $1\times1$ convolution on the last feature map.
By penalizing image content at the scale of patches, the color fidelity of synthesized images is guaranteed as illustrated in our ablation study (see Fig.~\ref{fig:syn_ablation}).
To align the class-conditional fake and real data distributions in the adversarial setting, the discriminator directly incorporates class labels as additional inputs in the manner of projection discriminator~\cite{miyato2018cgans}.
The class label is projected to a learned 512-dimensional class embedding and takes inner-product at every spatial position of the $8\times8\times512$ feature map.
The resulting $8\times8$ feature map is then added to the aforementioned $8\times8$ logit map, composing the final output of the discriminator.

For the objective function, we use the \textit{hinge} version~\cite{lim2017geometric} of the standard adversarial loss $\mathcal{L}_{adv}$. We also employ $R_1$ regularization $\mathcal{L}_{reg}$~\cite{mescheder2018training} as a slight gradient penalty for the discriminator.
Combining all the loss functions above, the total objective $\mathcal{L}_{total}$ to train the proposed CellGAN in an adversarial manner can be expressed as:

\begin{equation}
    \mathcal{L}_{total}=\mathcal{L}_{adv}+\mathcal{L}_{recon}+\lambda_{reg}\mathcal{L}_{reg},
\end{equation}

\noindent where $\lambda_{reg}$ is empirically set to 0.01 in our experiments.



\section{Experimental Results}
\subsection{Dataset and Experimental Setup}
\subsubsection{Dataset}
In this study, we collect 14,477 images with $256\times256$ pixels from three collaborative clinical centers. 
All the images are manually inspected to contain different cervical squamous cell types. 
In total, there are 7,662 NILM, 2,275 ASC-US, 2,480 LSIL, 1,638 ASC-H, and 422 HSIL images.
All the $256\times256$ images with their class labels are selected as the training data.

\subsubsection{Implementation Details}
We use the learning rate of $2.5\times10^{-4}$, batch size of 64, and Adam optimizer~\cite{kingma2014adam} to train both the generator and the discriminator for $100k$ iterations.
Spectral normalization~\cite{miyato2018spectral}, differentiable augmentation~\cite{zhao2020differentiable} and exponential-moving-average optimization~\cite{yazici2019unusual} are included in the training process.
Fr\'echet Inception Distance (FID)~\cite{heusel2017gans} is used to measure the overall semantic realism of the synthesized images.
All the experiments are conducted using an NVIDIA GeForce RTX 3090 GPU with PyTorch~\cite{paszke2019pytorch}.

\begin{figure}[ht]
    \centering
    \includegraphics[width=\textwidth]{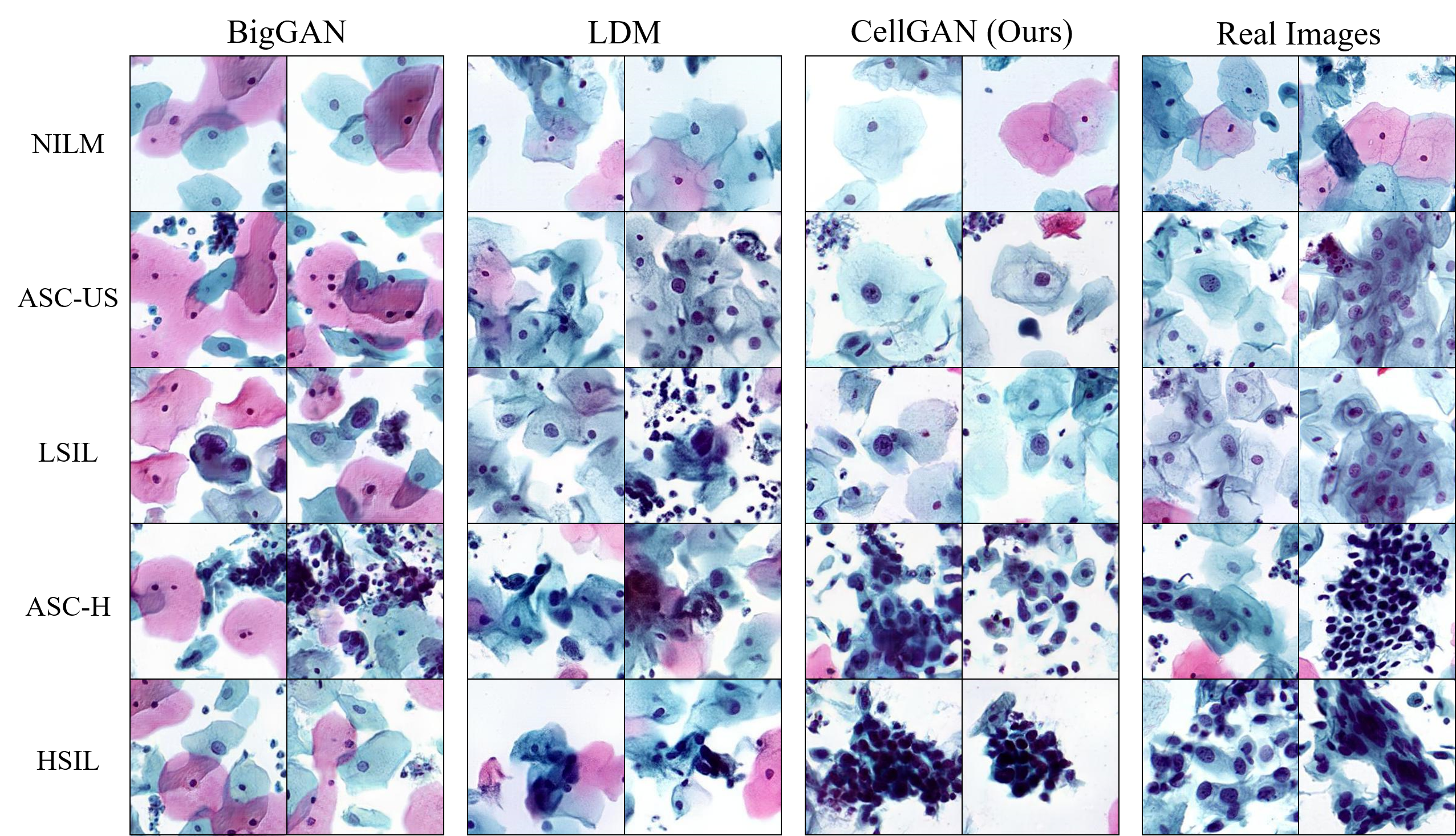}
    \caption{
    Qualitative comparison between state-of-the-art generative models and the proposed CellGAN. Different rows stand for different cervical squamous cell types.}
    \label{fig:syn_compare}
\end{figure}

\begin{table}[htb]
\centering
\caption{Quantitative comparison between state-of-the-art generative models and the proposed CellGAN (↓: Lower is better).}
\label{tab:syn_compare}
\setlength{\tabcolsep}{2mm}{
\begin{tabular}{c|cccccc}
\hline
\multirow{2}{*}{Method} & \multicolumn{6}{c}{FID↓} \\ \cline{2-7} 
 & NILM & ASC-US & LSIL & ASC-H & HSIL & Mean \\ \hline
BigGAN & 29.5076 & 37.9543 & 35.5058 & 48.0228 & 85.6230 & 47.3227 \\
LDM & 53.4307 & 56.1689 & 49.0969 & 59.6406 & 84.9522 & 60.6579 \\ 
CellGAN(Ours) & \textbf{26.0135} & \textbf{33.5718} & \textbf{33.3401} & \textbf{46.2965} & \textbf{68.3458} & \textbf{41.5136} \\ \hline
\end{tabular}
}
\end{table}

\subsection{Evaluation of Image Synthesis Quality}
We compare CellGAN with the state-of-the-art generative models for class-conditional image synthesis, \ie, BigGAN~\cite{brock2018large} from cGANs~\cite{mirza2014conditional} and Latent Diffusion Model (LDM)~\cite{rombach2022high} from diffusion models~\cite{ho2020denoising}.
As shown in Fig.~\ref{fig:syn_compare}, BigGAN cannot generate individual cells with clearly defined cell boundaries.
And it also fails to capture the morphological features of HSIL cells that are relatively limited in training data quantity.
LDM only yields half-baked cell structures since the generated cells are mixed, and there exists negligible class separability among abnormal cell types. 
On the contrary, our proposed CellGAN is able to synthesize visually plausible cervical cells and accurately model distinguishable cellular features for each cell type. 
The quantitative comparison by FID in Table~\ref{tab:syn_compare} also demonstrates the superiority of CellGAN in synthesized image quality.

\begin{figure}[ht]
    \centering
    \includegraphics[width=\textwidth]{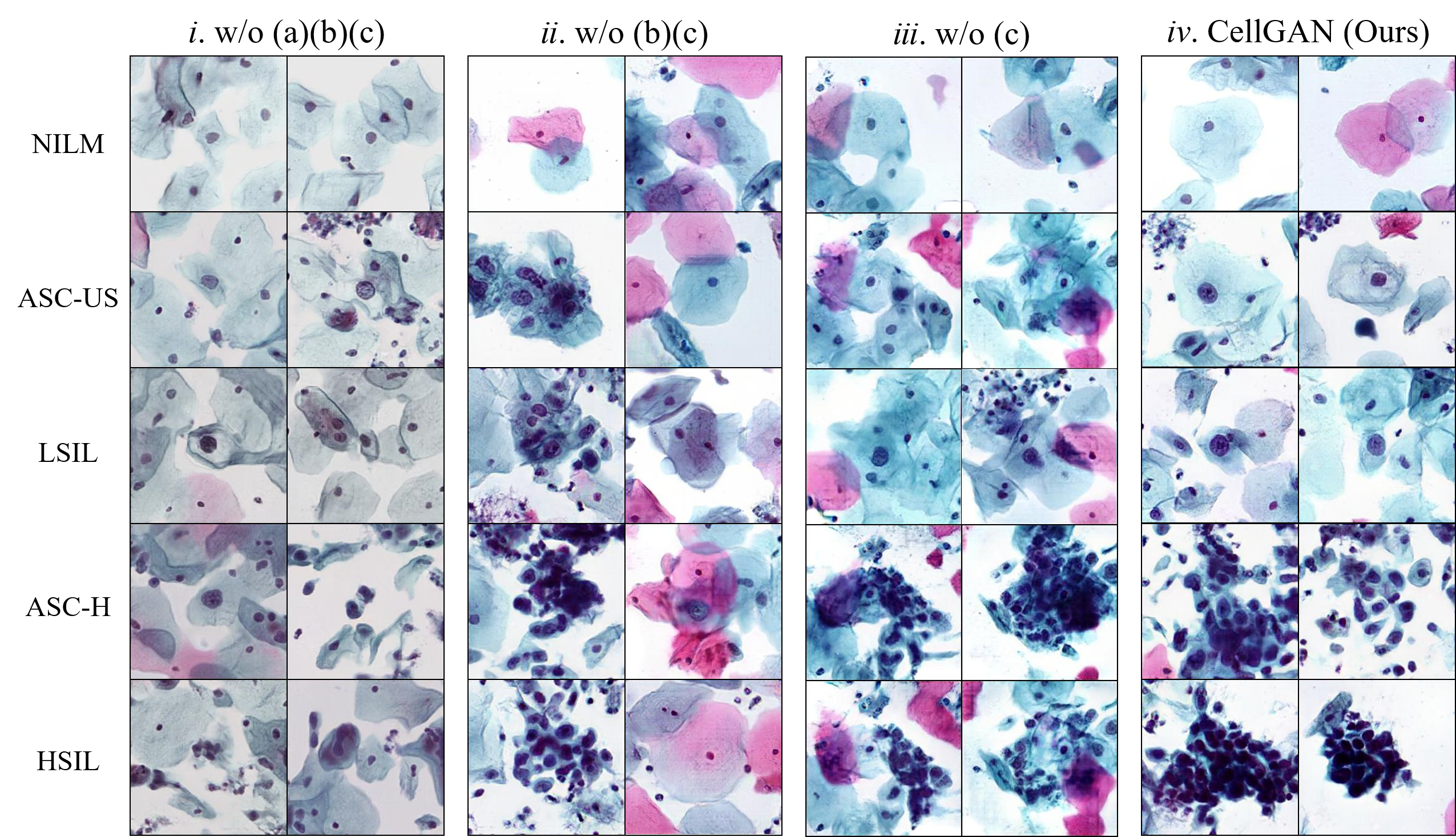}
    \caption{Generated images from ablation study of the following key components: (a) PatchGAN architecture, (b) class mapping network, (c) SGC module.}
    \label{fig:syn_ablation}
\end{figure}

To verify the effects of key components in the proposed CellGAN, we conduct an ablation study on four model settings in Table~\ref{tab:syn_ablation} and Fig.~\ref{fig:syn_ablation}.
We denote the models in Fig.~\ref{fig:syn_ablation} from left to right as \textit{Model i}, \textit{Model ii}, \textit{Model iii}, and CellGAN.
The visual results of \textit{Model i} suffer from severe color distortions while the other models do not, indicating that the PatchGAN-based discriminator can guarantee color fidelity by patch-level image content penalty.
\begin{table}[htb]
\centering
\caption{Quantitative ablation study of the following key component: (a) PatchGAN architecture, (b) class mapping network, (c) SGC module. (↓: Lower is better).}
\label{tab:syn_ablation}
\setlength{\tabcolsep}{2mm}{
\begin{tabular}{ccc|cccccc}
\hline
\multicolumn{3}{c|}{Model Setting} & \multicolumn{6}{c}{FID↓} \\ \hline
(a) & (b) & (c) & NILM & ASC-US & LSIL & ASC-H & HSIL & Mean \\ \hline
- & - & - & 39.7048 & 45.9424 & 42.2336 & 58.3448 & 84.0265 & 54.0504 \\
$\surd$ & - & - & 42.7974 & 36.1800 & 38.1507 & 52.0282 & 74.4304 & 48.7173 \\
$\surd$ & $\surd$ & - & 31.1720 & 38.1468 & 39.5540 & 47.2584 & 68.6068 & 44.9476 \\
$\surd$ & $\surd$ & $\surd$ & \textbf{26.0135} & \textbf{33.5718} & \textbf{33.3401} & \textbf{46.2965} & \textbf{68.3458} & \textbf{41.5136} \\ \hline
\end{tabular}
}
\end{table}

The abnormal cells generated by \textit{Model i} and \textit{Model ii} tend to have highly similar cellular features.
In contrast, \textit{Model iii} and CellGAN can accurately capture the morphological characteristics of different cell types.
This phenomenon suggests that the implementation of the class mapping network facilitates more distinguishable feature representations for different cell types.
By comparing the synthesized images from \textit{Model iii} with CellGAN, it is observed that adopting SGC modules can yield more clear cell boundaries, which demonstrates the capability of SGC module in modeling complicated cell-to-cell relationships in image space.
The quantitative results further state the effects of the components above. 

\begin{table}[htb]
\centering
\caption{Data augmentation comparison between the proposed CellGAN and other synthesis-based methods (↑: Higher is better).}
\label{tab:da_compare}
\setlength{\tabcolsep}{2mm}{
\begin{tabular}{c|c|cccc}
\hline
Classifier & Method & Accuracy↑ & Precision↑ & Recall↑ & F1-Score↑ \\ \hline
\multirow{4}{*}{ResNet} & baseline & $74.30_{\pm1.69}$ & $68.00_{\pm1.79}$ & $70.94_{\pm2.28}$ & $68.88_{\pm1.78}$ \\
 & + BigGAN & $76.30_{\pm2.80}$ & $72.96_{\pm2.58}$ & $75.11_{\pm2.20}$ & $73.89_{\pm2.44}$ \\
 & + LDM & $75.80_{\pm1.12}$ & $71.14_{\pm0.72}$ & $73.89_{\pm1.36}$ & $72.29_{\pm0.91}$ \\
 & + CellGAN & $\textbf{79.55}_{\pm\textbf{1.20}}$ & $\textbf{74.88}_{\pm\textbf{1.60}}$ & $\textbf{75.42}_{\pm\textbf{1.74}}$ & $\textbf{74.70}_{\pm\textbf{1.79}}$ \\ \hline
\multirow{4}{*}{DenseNet} & baseline & $72.10_{\pm0.66}$ & $65.23_{\pm1.17}$ & $68.28_{\pm1.26}$ & $66.33_{\pm1.25}$ \\
 & + BigGAN & $75.40_{\pm1.73}$ & $68.47_{\pm1.78}$ & $70.13_{\pm1.21}$ & $68.94_{\pm1.97}$ \\
 & + LDM & $74.95_{\pm1.94}$ & $68.03_{\pm2.11}$ & $69.32_{\pm1.65}$ & $68.55_{\pm2.37}$ \\
 & + CellGAN & $\textbf{76.15}_{\pm\textbf{1.38}}$ & $\textbf{70.37}_{\pm\textbf{1.52}}$ & $\textbf{72.42}_{\pm\textbf{1.95}}$ & $\textbf{70.99}_{\pm\textbf{1.75}}$ \\ \hline
\end{tabular}
}
\end{table}

\subsection{Evaluation of Augmentation Effectiveness}
To validate the data augmentation capacity of the proposed CellGAN, we conduct 5-fold cross-validations on the cell classification performances of two classifiers (ResNet-34~\cite{he2016deep} and DenseNet-121~\cite{huang2017densely}) using four training data settings for comparison:
(1) real data only (the baseline); 
(2) baseline + BigGAN synthesized images; 
(3) baseline + LDM synthesized images; 
(4) baseline + CellGAN synthesized images. 
For each cell type, we randomly select 400 real images and divide them into 5 groups.
In each fold, one group is selected as the testing data while the other four are used for training.
For different data settings, we synthesize 2,000 images for each cell type using the corresponding generative method, and add them to the training data of each fold.
We use the learning rate of $1.0\times10^{-4}$, batch size of 64, and SGD optimizer~\cite{robbins1951stochastic} to train all the classifiers for 30 epochs.
Random flip is applied to all data settings since it is reasonable to use traditional data augmentation techniques simultaneously in practice.

The experimental accuracy, precision, recall, and F1 score are listed in Table~\ref{tab:da_compare}. 
It is shown that both the classifiers achieve the best scores in all metrics using the additional synthesized data from CellGAN. 
Compared with the baselines, the accuracy values of ResNet-34 and DenseNet-121 are improved by 5.25\% and 4.05\%, respectively.
Meanwhile, the scores of other metrics are all improved by more than 4\%, indicating that our synthesized data can significantly enhance the overall classification performance.
Thanks to the visually plausible and semantically realistic synthesized data, CellGAN is conducive to the improvement of cell classification, thus serving as an efficient tool for augmenting automatic abnormal cervical cell screening.

\section{Conclusion and Discussion}
In this paper, we propose CellGAN for class-conditional cytopathological image synthesis of different cervical cell types.
Built upon FastGAN for training stability and computational efficiency, incorporating class-conditional information of cell types via non-linear mapping can better represent distinguishable cellular features.
The proposed SGC module provides the global contexts of cell spatial relationships by capturing long-range dependencies.
We have also found that the PatchGAN-based discriminator can prevent potential color distortion.
Qualitative and quantitative experiments validate the semantic realism as well as the data augmentation effectiveness of the synthesized images from CellGAN.

Meanwhile, our current CellGAN still has several limitations.
First, we cannot explicitly control the detailed attributes of the synthesized cell type, \eg, nucleus size, and nucleus-cytoplasm ratio.
Second, in this paper, the synthesized image size is limited to $256\times256$.
It is worth conducting more studies for expanding synthesized image size to contain much more cells, such that the potential applications can be extended to other clinical scenes (\eg, interactively training pathologists) in the future.

\subsubsection{Acknowledgement.} This work was supported by the National Natural Science Foundation of China (No. 62001292).

%
%
\bibliographystyle{splncs04}
\bibliography{ref.bib}

\begin{thebibliography}{10}
\providecommand{\url}[1]{\texttt{#1}}
\providecommand{\urlprefix}{URL }
\providecommand{\doi}[1]{https://doi.org/#1}

\bibitem{abulafia2003performance}
Abulafia, O., Pezzullo, J.C., Sherer, D.M.: Performance of thinprep
  liquid-based cervical cytology in comparison with conventionally prepared
  papanicolaou smears: a quantitative survey. Gynecologic oncology
  \textbf{90}(1),  137--144 (2003)

\bibitem{brock2018large}
Brock, A., Donahue, J., Simonyan, K.: Large scale gan training for high
  fidelity natural image synthesis. In: International Conference on Learning
  Representations (2018)

\bibitem{cao2021novel}
Cao, L., Yang, J., Rong, Z., Li, L., Xia, B., You, C., Lou, G., Jiang, L., Du,
  C., Meng, H., et~al.: A novel attention-guided convolutional network for the
  detection of abnormal cervical cells in cervical cancer screening. Medical
  image analysis  \textbf{73},  102197 (2021)

\bibitem{cao2019gcnet}
Cao, Y., Xu, J., Lin, S., Wei, F., Hu, H.: Gcnet: Non-local networks meet
  squeeze-excitation networks and beyond. In: Proceedings of the IEEE/CVF
  international conference on computer vision workshops. pp.~0--0 (2019)

\bibitem{davey1994atypical}
Davey, D.D., Naryshkin, S., Nielsen, M.L., Kline, T.S.: Atypical squamous cells
  of undetermined significance: interlaboratory comparison and quality
  assurance monitors. Diagnostic cytopathology  \textbf{11}(4),  390--396
  (1994)

\bibitem{gultekin2020world}
Gultekin, M., Ramirez, P.T., Broutet, N., Hutubessy, R.: World health
  organization call for action to eliminate cervical cancer globally.
  International Journal of Gynecological Cancer  \textbf{30}(4),  426--427
  (2020)

\bibitem{he2016deep}
He, K., Zhang, X., Ren, S., Sun, J.: Deep residual learning for image
  recognition. In: Proceedings of the IEEE conference on computer vision and
  pattern recognition. pp. 770--778 (2016)

\bibitem{heusel2017gans}
Heusel, M., Ramsauer, H., Unterthiner, T., Nessler, B., Hochreiter, S.: Gans
  trained by a two time-scale update rule converge to a local nash equilibrium.
  Advances in neural information processing systems  \textbf{30} (2017)

\bibitem{ho2020denoising}
Ho, J., Jain, A., Abbeel, P.: Denoising diffusion probabilistic models.
  Advances in Neural Information Processing Systems  \textbf{33},  6840--6851
  (2020)

\bibitem{hou2019robust}
Hou, L., Agarwal, A., Samaras, D., Kurc, T.M., Gupta, R.R., Saltz, J.H.: Robust
  histopathology image analysis: To label or to synthesize? In: Proceedings of
  the IEEE/CVF Conference on Computer Vision and Pattern Recognition. pp.
  8533--8542 (2019)

\bibitem{huang2017densely}
Huang, G., Liu, Z., Van Der~Maaten, L., Weinberger, K.Q.: Densely connected
  convolutional networks. In: Proceedings of the IEEE conference on computer
  vision and pattern recognition. pp. 4700--4708 (2017)

\bibitem{isola2017image}
Isola, P., Zhu, J.Y., Zhou, T., Efros, A.A.: Image-to-image translation with
  conditional adversarial networks. In: Proceedings of the IEEE conference on
  computer vision and pattern recognition. pp. 1125--1134 (2017)

\bibitem{karras2019style}
Karras, T., Laine, S., Aila, T.: A style-based generator architecture for
  generative adversarial networks. In: Proceedings of the IEEE/CVF conference
  on computer vision and pattern recognition. pp. 4401--4410 (2019)

\bibitem{kingma2014adam}
Kingma, D.P., Ba, J.: Adam: A method for stochastic optimization. arXiv
  preprint arXiv:1412.6980  (2014)

\bibitem{lim2017geometric}
Lim, J.H., Ye, J.C.: Geometric gan. arXiv preprint arXiv:1705.02894  (2017)

\bibitem{liu2020towards}
Liu, B., Zhu, Y., Song, K., Elgammal, A.: Towards faster and stabilized gan
  training for high-fidelity few-shot image synthesis. In: International
  Conference on Learning Representations (2020)

\bibitem{mescheder2018training}
Mescheder, L., Geiger, A., Nowozin, S.: Which training methods for gans do
  actually converge? In: International conference on machine learning. pp.
  3481--3490. PMLR (2018)

\bibitem{mirza2014conditional}
Mirza, M., Osindero, S.: Conditional generative adversarial nets. arXiv
  preprint arXiv:1411.1784  (2014)

\bibitem{miyato2018spectral}
Miyato, T., Kataoka, T., Koyama, M., Yoshida, Y.: Spectral normalization for
  generative adversarial networks. In: International Conference on Learning
  Representations (2018)

\bibitem{miyato2018cgans}
Miyato, T., Koyama, M.: cgans with projection discriminator. In: International
  Conference on Learning Representations (2018)

\bibitem{nayar2015bethesda}
Nayar, R., Wilbur, D.C.: The Bethesda system for reporting cervical cytology:
  definitions, criteria, and explanatory notes. Springer (2015)

\bibitem{paszke2019pytorch}
Paszke, A., Gross, S., Massa, F., Lerer, A., Bradbury, J., Chanan, G., Killeen,
  T., Lin, Z., Gimelshein, N., Antiga, L., et~al.: Pytorch: An imperative
  style, high-performance deep learning library. Advances in neural information
  processing systems  \textbf{32} (2019)

\bibitem{patel2011cervical}
Patel, M.M., Pandya, A.N., Modi, J.: Cervical pap smear study and its utility
  in cancer screening, to specify the strategy for cervical cancer control.
  National journal of community medicine  \textbf{2}(01),  49--51 (2011)

\bibitem{robbins1951stochastic}
Robbins, H., Monro, S.: A stochastic approximation method. The annals of
  mathematical statistics pp. 400--407 (1951)

\bibitem{rombach2022high}
Rombach, R., Blattmann, A., Lorenz, D., Esser, P., Ommer, B.: High-resolution
  image synthesis with latent diffusion models. In: Proceedings of the IEEE/CVF
  Conference on Computer Vision and Pattern Recognition. pp. 10684--10695
  (2022)

\bibitem{shorten2019survey}
Shorten, C., Khoshgoftaar, T.M.: A survey on image data augmentation for deep
  learning. Journal of big data  \textbf{6}(1),  1--48 (2019)

\bibitem{xiang2020novel}
Xiang, Y., Sun, W., Pan, C., Yan, M., Yin, Z., Liang, Y.: A novel
  automation-assisted cervical cancer reading method based on convolutional
  neural network. Biocybernetics and Biomedical Engineering  \textbf{40}(2),
  611--623 (2020)

\bibitem{xue2021selective}
Xue, Y., Ye, J., Zhou, Q., Long, L.R., Antani, S., Xue, Z., Cornwell, C.,
  Zaino, R., Cheng, K.C., Huang, X.: Selective synthetic augmentation with
  histogan for improved histopathology image classification. Medical image
  analysis  \textbf{67},  101816 (2021)

\bibitem{yazici2019unusual}
Yazici, Y., Foo, C.S., Winkler, S., Yap, K.H., Piliouras, G., Chandrasekhar,
  V., et~al.: The unusual effectiveness of averaging in gan training. In: ICLR
  (Poster) (2019)

\bibitem{zhao2020differentiable}
Zhao, S., Liu, Z., Lin, J., Zhu, J.Y., Han, S.: Differentiable augmentation for
  data-efficient gan training. Advances in Neural Information Processing
  Systems  \textbf{33},  7559--7570 (2020)

\bibitem{zhou2021hierarchical}
Zhou, M., Zhang, L., Du, X., Ouyang, X., Zhang, X., Shen, Q., Luo, D., Fan, X.,
  Wang, Q.: Hierarchical pathology screening for cervical abnormality.
  Computerized Medical Imaging and Graphics  \textbf{89},  101892 (2021)

\end{thebibliography}

\title{\makebox[\linewidth]{\parbox{\dimexpr\textwidth+2cm\relax}{\centering Supplementary Materials for Paper Titled \\ "CellGAN: Conditional Cervical Cell Synthesis for \\ Augmenting Cytopathological Image Classification"}}}

\author{{\makebox[\linewidth]{\parbox{\dimexpr\textwidth+2cm\relax}{\centering
Zhenrong Shen\inst{1} \and
Maosong Cao\inst{2} \and
Sheng Wang\inst{1,3} \and
Lichi Zhang\inst{1} \and
Qian Wang\inst{2}$^{(\textrm{\Letter})}$
}}}}
\authorrunning{Z. Shen et al.}

\institute{
School of Biomedical Engineering, Shanghai Jiao Tong University, Shanghai, China \and
School of Biomedical Engineering, ShanghaiTech University, Shanghai, China
\email{wangqian2@shanghaitech.edu.cn} \and
Shanghai United Imaging Intelligence Co., Ltd., Shanghai, China
}

\maketitle  

\begin{figure}[htbp]
    \centering
    \includegraphics[width=\textwidth]{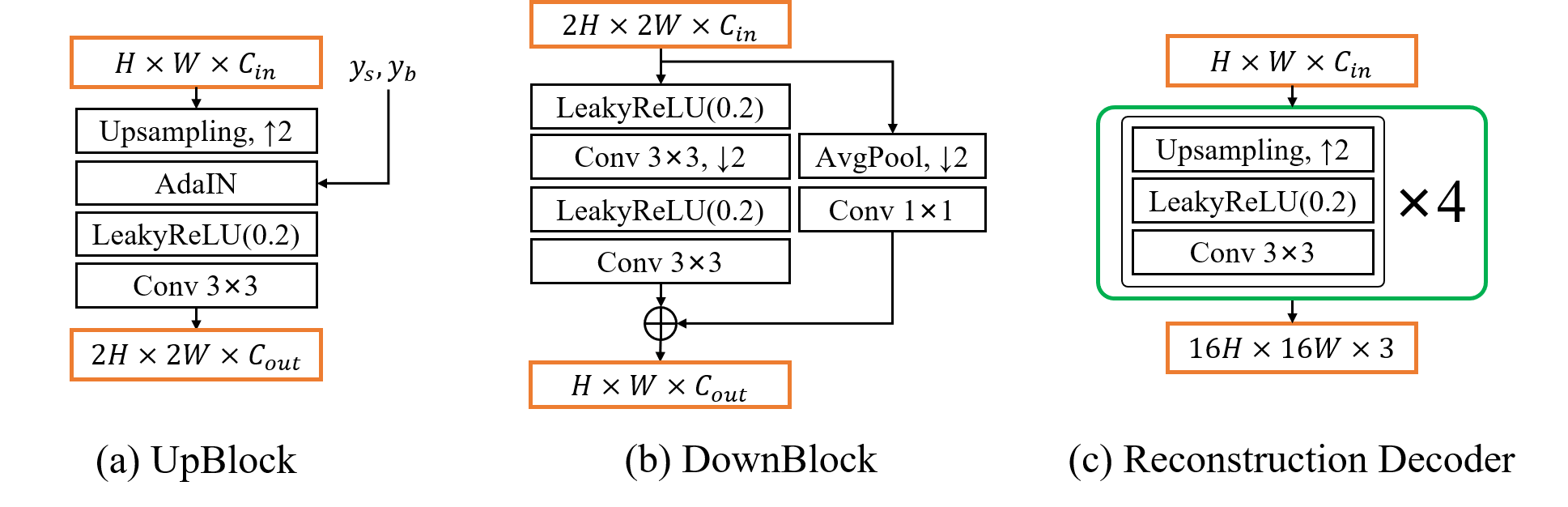}
    \caption{(a) UpBlock in the generator, where $y_{s}$ and $y_{b}$ stand for the scaling and bias parameters of AdaIN, respectively. (b) DownBlock in the discriminator. (c) Reconstruction Decoder in the discriminator for self-supervised regularization.}
    \label{fig:overview}
\end{figure}

\vspace{-0.8cm}

\begin{figure}[htbp]
    \centering
    \includegraphics[width=\textwidth]{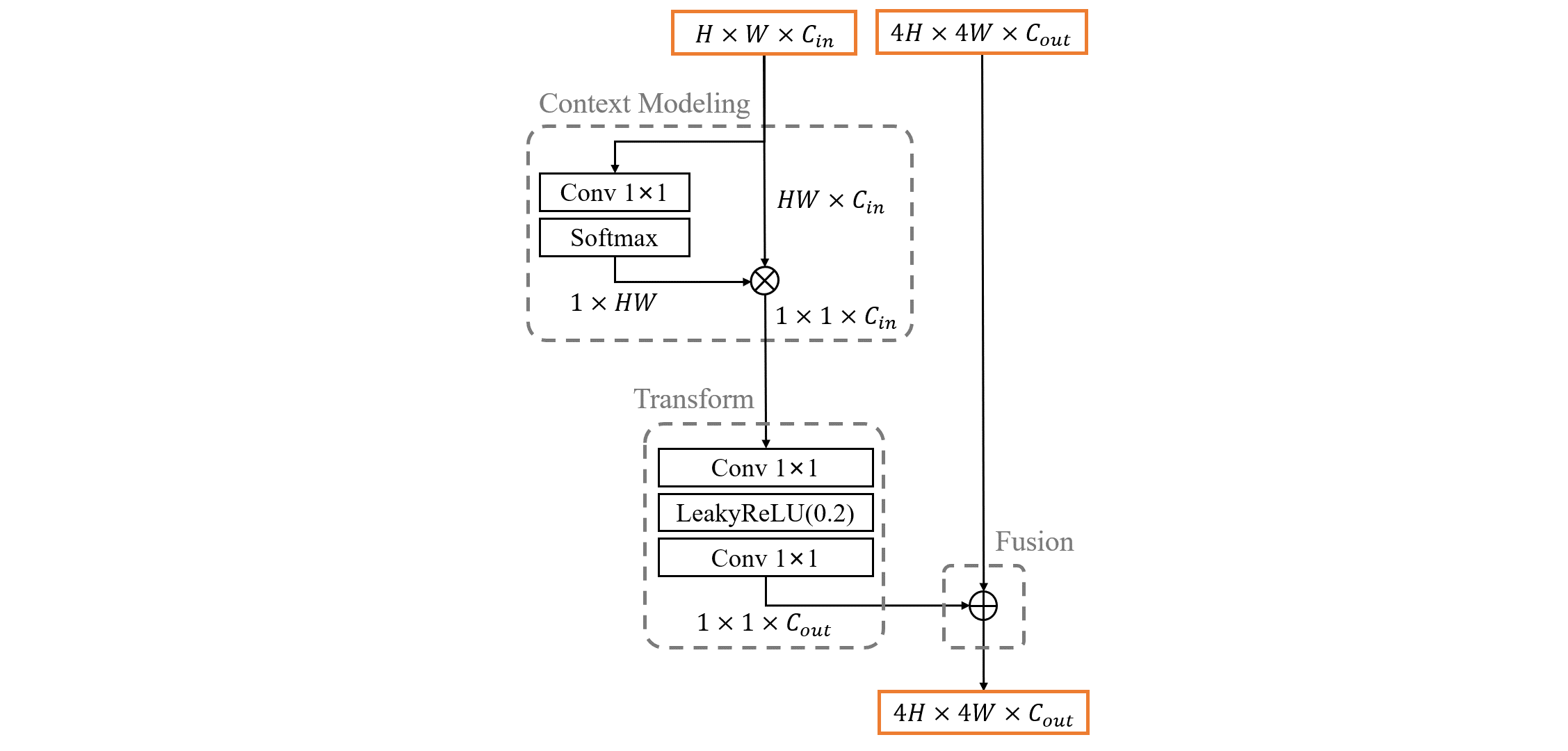}
    \caption{Skip-layer Global Context (SGC) Module in the generator.}
    \label{fig:overview}
\end{figure}

\end{document}